\documentclass[10pt,twocolumn,letterpaper]{article}

\usepackage[pagenumbers]{wacv}

\usepackage[table]{xcolor}

\definecolor{pearFour}{HTML}{588F27}
\definecolor{pearFith}{HTML}{ECF0F1}
\definecolor{pearDark}{HTML}{2980B9}
\definecolor{pearDarker}{HTML}{1D2DEC}
\definecolor{pearLight}{HTML}{F7E0D5}

\definecolor{eva01purple}{RGB}{168,119,200}
\definecolor{eva02red}{RGB}{236,35,35}

\usepackage{url}
\usepackage{nicefrac}
\usepackage{lipsum}
\usepackage{fancyhdr}
\usepackage{appendix}
\usepackage{afterpage}

\usepackage{graphicx}
\usepackage{caption}
\usepackage{subcaption}
\usepackage{wrapfig}
\usepackage{float}

\usepackage{booktabs}
\usepackage{multirow}
\usepackage{threeparttable}
\usepackage{arydshln}

\usepackage{amsmath}
\usepackage{amssymb}
\usepackage{amsfonts}
\usepackage{bm}
\usepackage{dsfont}
\DeclareMathOperator*{\argmin}{arg\,min}

\usepackage{algorithm}
\usepackage{algorithmic}
\usepackage{listings}
\usepackage{verbatim}
\usepackage{comment}

\usepackage{etoolbox}
\makeatletter
\AfterEndEnvironment{algorithm}{\let\@algcomment\relax}
\AtEndEnvironment{algorithm}{\kern2pt\hrule\relax\vskip3pt\@algcomment}
\let\@algcomment\relax
\newcommand\algcomment[1]{\def\@algcomment{\footnotesize#1}}
\renewcommand\fs@ruled{\def\@fs@cfont{\bfseries}\let\@fs@capt\floatc@ruled
  \def\@fs@pre{\hrule height.8pt depth0pt \kern2pt}%
  \def\@fs@post{}%
  \def\@fs@mid{\kern2pt\hrule\kern2pt}%
  \let\@fs@iftopcapt\iftrue}
\makeatother

\usepackage{pifont}

\usepackage{textcomp}
\usepackage{relsize}
\usepackage{xspace}

\usepackage{cancel}
\newcommand{\redcancelto}[2]{%
    \textcolor{red}{\cancelto{#1}{#2}}%
}

\usepackage{tikz}
\usetikzlibrary{decorations.pathreplacing, positioning}

\definecolor{wacvblue}{rgb}{0.21,0.49,0.74}
\usepackage[pagebackref,breaklinks,colorlinks,allcolors=wacvblue]{hyperref}

\def\wacvPaperID{*****}
\def\confName{WACV}
\def\confYear{2027}

\title{
Relational Representation Distillation
}

\author{
Nikos Giakoumoglou \qquad Tania Stathaki\\
Imperial College London\\
{\tt\small \{nikos,tania\}@imperial.ac.uk}\vspace{.5em}\\
{Code: \url{https://github.com/giakoumoglou/rrd}}\\
}

\begin{document}
\maketitle

\begin{abstract}
Knowledge distillation transfers knowledge from large teacher models to more compact student networks. The standard approach minimizes the Kullback--Leibler (KL) divergence between the probabilistic outputs of the teacher and student, aligning predictions but neglecting the structural relationships encoded within the teacher's internal representations. Recent advances have adopted contrastive learning objectives to address this limitation; however, such instance-discrimination--based methods induce a \textit{``class collision problem''}, in which semantically related samples are inappropriately pushed apart despite belonging to similar classes. To overcome this, we propose \textbf{\underline{R}}elational \textbf{\underline{R}}epresentation \textbf{\underline{D}}istillation (RRD) that preserves the \textit{relative relationships} among instances rather than enforcing absolute separation. Our method introduces separate temperature parameters for teacher and student distributions, with a \textit{sharper teacher} (low~$\tau_t$) emphasizing primary relationships and a \textit{softer student} (high~$\tau_s$) maintaining secondary similarities. This dual-temperature formulation creates an implicit information bottleneck that preserves fine-grained relational structure while avoiding the over-separation characteristic of contrastive losses. We establish theoretical connections showing that InfoNCE emerges as a limiting case of our objective when~$\tau_t \rightarrow 0$, and empirically demonstrate that this relaxed formulation yields superior relational alignment and generalization across classification and detection tasks.
\end{abstract}

\section{Introduction}
\label{sec:introduction}

\begin{figure}[!t]
    \includegraphics[width=0.95\linewidth]{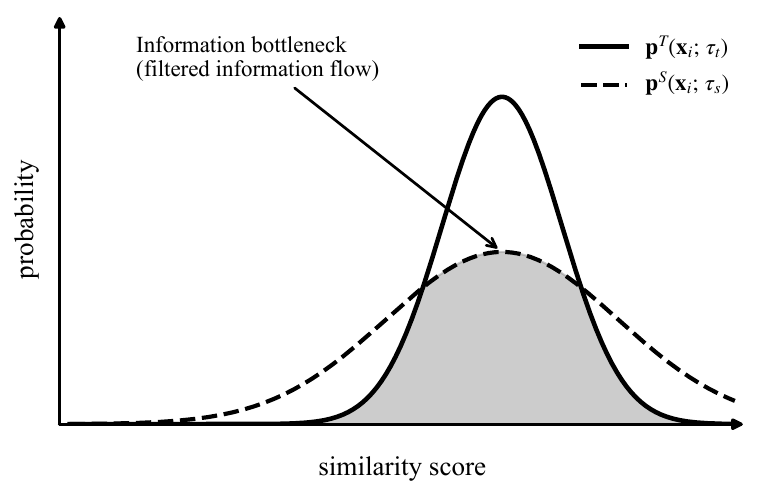}
    \caption{\textbf{Visualization of the \textit{information bottleneck} effect.} The teacher produces a sharper similarity distribution $\mathbf{p}^T(\mathbf{x}_i;\tau_t)$ (solid black) highlighting primary relationships, while the student adopts a softer distribution $\mathbf{p}^S(\mathbf{x}_i;\tau_s)$ (dashed black) that retains secondary similarities. The gray-shaded overlap region illustrates the \textit{filtered information flow}, where only essential relational cues are transferred from teacher to student, effectively bounding $I(\mathbf{z}^T;\mathbf{z}^S)$.}
    \label{fig:info_bottleneck}
\end{figure}

Knowledge Distillation (KD) transfers knowledge from large, high-capacity \textit{teacher} models to compact \textit{student} models~\citep{hinton2015distilling}. This approach is increasingly relevant as vision models for image classification~\citep{liu2021swin, dosovitskiy2021vit}, object detection~\citep{ren2016fasterrcnn, lin2017feature}, and semantic segmentation~\citep{chen2017deeplab, chen2017deeplabv3} continue to grow in size and computational cost~\citep{goyal2019scaling, kornblith2019do}, motivating efficient model compression techniques~\citep{bucilua2006model, polino2018model}.

The seminal work of \citet{bucilua2006model} and \citet{hinton2015distilling} introduced the idea of transferring knowledge by minimizing the \textit{Kullback--Leibler} (KL) divergence between teacher and student output distributions. This formulation makes intuitive sense when the output is a categorical probability mass function over classes. However, in many cases, we wish to transfer richer internal knowledge — not just about class probabilities but about the underlying \textit{representations} that encode visual semantics and inter-class relations. Subsequent works have emphasized the importance of intermediate feature representations~\citep{romero2014fitnets,zagoruyko2016paying,tung2019similarity,park2019relational}.

Representational knowledge is inherently \textit{structured}: feature dimensions exhibit non-trivial correlations and higher-order dependencies. Logit matching alone cannot capture this relational structure. To address this, feature-based methods~\citep{romero2014fitnets,zagoruyko2016paying,yim2017gift,peng2019correlation} extend distillation to intermediate representations. However, \citet{tian2022crd} showed that such approaches still neglect the structural knowledge encoded in the teacher's internal representations. To overcome this limitation, \citet{tian2022crd} further adapted the family of contrastive objectives~\citep{gutmann2010noise,oord2019cpc,arora2019theoretical,hjelm2019dim} for distilling structured knowledge between teacher and student networks. These objectives have been highly successful in density estimation and self-supervised representation learning, as they implicitly maximize a lower bound on the mutual information between paired embeddings.

However, such instance-discrimination--based approaches inevitably introduce a \textit{``class-collision problem''}~\citep{arora2019theoretical,li2021pcl,yeh2022dcl}, in which semantically similar samples are undesirably pushed apart due to uniform negative sampling. As shown in our experiments (\Cref{tab:semantic}), this repulsion degrades semantic coherence, leading to fragmented feature spaces where related instances lose proximity. 

To overcome these limitations, we propose a method that relaxes rigid contrastive objectives by preserving meaningful \textit{relative relationships} between instances in feature space, which we term \textbf{\underline{R}}elational \textbf{\underline{R}}epresentation \textbf{\underline{D}}istillation (RRD). For example, given images of a \textit{``cat''}, \textit{``dog''}, and \textit{``plane''}, what matters is not absolute similarity scores but their relative ordering: the cat should be closest to another cat, followed by the dog (another animal), and farthest from the plane. RRD achieves this by aligning the teacher and student similarity distributions through a KL-based loss, rather than enforcing one-hot positive matches. We introduce distinct temperature parameters for the teacher and student distributions: a \textbf{sharper teacher} (low~$\tau_t$) emphasizes primary relationships, while a \textbf{softer student} (high~$\tau_s$) retains secondary similarities. This dual-temperature mechanism naturally forms an \textit{information bottleneck} (\Cref{fig:info_bottleneck}), bounding the information transferred between teacher and student to only the most salient relational cues. As shown empirically, this formulation mitigates class collisions and yields superior structural alignment compared to prior methods such as CRD~\citep{tian2022crd}.

Our main \textbf{contributions} are as follows: \textbf{(i)} We propose an objective that preserves structural relationships between feature representations using distinct temperature parameters for the teacher and student, forming an implicit information bottleneck that balances sharp primary alignment with smooth secondary similarities. \textbf{(ii)} We establish theoretical connections between our objective, InfoNCE~\citep{oord2019cpc}, and the KL divergence, showing that InfoNCE arises as a limiting case when $\tau_t \rightarrow 0$. \textbf{(iii)} We empirically demonstrate the advantages of our objective across classification (\Cref{tab:cifar100,tab:imagenet}) and detection (\Cref{tab:coco}) benchmarks, achieving consistent gains over existing methods. \textbf{(iv)} We provide quantitative and qualitative analyses of learned representations through correlation alignment (\Cref{tab:corralign}), and semantic similarity evaluation (\Cref{tab:semantic}) confirming that RRD maintains relational topology between teacher and student embeddings.

\section{Related Work}
\label{sec:related_work}

The seminal works of \citet{bucilua2006model} and \citet{hinton2015distilling} established the foundation of knowledge distillation, where compact student networks learn from large teacher models with minimal loss in generalization. Later extensions such as \citet{li2014learning} refined this formulation for better transferability. Since then, knowledge distillation has evolved into several branches, most notably \textit{logit-based}~\citep{mirzadeh2019takd, yang2019learning, huang2022dist} and \textit{feature-based}~\citep{romero2014fitnets, zagoruyko2016paying, yim2017gift, tian2022crd} distillation.

\subsection{Logit-based Distillation}
\label{subsec:logit_based}

Logit-based approaches transfer knowledge by matching the \textit{output logits} (the pre-softmax scores) of the teacher and student, encouraging the student to mimic the teacher's predictive distribution and class-level semantics. Early methods improved stability and transfer via hierarchical supervision~\citep{yang2019learning}, multi-step training~\citep{mirzadeh2019takd}, or collaborative learning~\citep{zhang2017deep}. Subsequent work refined this process by adjusting how logits are represented or weighted~\citep{zhou2021wsld,yuan2020revisiting,niu2022ipwd,sun2024logitstandardization}. Further refinements involved dynamic temperature adjustment~\citep{li2022ctkd}, transformation-based alignment~\citep{zheng2024ttm}, and adaptive teacher calibration~\citep{huang2022dist}.

\subsection{Feature-based Distillation}
\label{subsec:feature_based}

Feature-based methods transfer richer structural knowledge by aligning intermediate teacher and student representations to capture spatial or semantic relationships. Foundational work explored hint-based supervision~\citep{romero2014fitnets}, attention transfer~\citep{zagoruyko2016paying}, and feature transformation~\citep{kim2018paraphrasing}, later extended by functional consistency~\citep{liu2023fcfd}, class-level attention~\citep{guo2023cat}, and structural normalization~\citep{chen2021reviewkd, chen2022simkd, liu2023norm}. A major branch focuses on preserving \textit{relational structures} among embeddings. Early works modeled such relations through inner products~\citep{yim2017gift}, distance preservation~\citep{park2019relational}, or correlation congruence~\citep{peng2019correlation}, while contrastive methods~\citep{tian2022crd} maximized mutual information via memory banks. Recent studies introduced redundancy reduction and kernel-based alignment~\citep{miles2024vkd, he2022fkd}. Our objective fits within this broader class of \textit{feature-relational distillation} methods but relaxes hard contrastive constraints, focusing instead on smooth relational distributions that preserve semantic coherence.

\subsection{Connection to Information-theoretic Objectives}
\label{subsec:info_theoretic}

Our objective is also related to \textit{InfoNCE}~\citep{oord2019cpc} and \textit{Noise-Contrastive Estimation (NCE)}~\citep{gutmann2010noise}, which maximize a lower bound on mutual information between representations~\citep{hjelm2019dim}. In this context, our dual-temperature formulation can be viewed as an information bottleneck that bounds the mutual information \(I(\mathbf{z}^T;\mathbf{z}^S)\), ensuring that only essential relational cues are transferred during distillation (\Cref{fig:info_bottleneck}).

\section{Methodology}
\label{sec:methodology}

Here, we introduce our objective which transfers knowledge from a pre-trained teacher network to a student network by leveraging relational cues embedded in their feature representations. \Cref{subsec:preliminaries} outlines the fundamental principles of knowledge distillation, \Cref{subsec:rrd} details the formulation of our relational objective, and \Cref{subsec:info_bottleneck} provides an analytical interpretation of the \textit{information bottleneck} that regulates the flow of relational knowledge.

\subsection{Preliminaries on Knowledge Distillation}
\label{subsec:preliminaries}

Knowledge distillation transfers knowledge from a high-capacity teacher network \(f^T_\theta\) to a compact student network \(f^S_\theta\)~\citep{hinton2015distilling,bucilua2006model}. Its primary objective is to enable the student model to approximate the performance of the teacher model while leveraging the student's computational efficiency. The overall distillation process can be formulated as:

\begin{equation}
    \hat{\theta}_{S} = \argmin_{\theta_{S}}\sum_{i}^{N} \left(
    \mathcal{L}_{\text{sup}}(\mathbf{x}_i, \theta_{S}, y_i) +
    \mathcal{L}_{{\text{distill}}}(\mathbf{x}_i, \theta_{S},  \theta_{T}) \right),
    \label{eq:supervised_distill}
\end{equation}

\noindent where $\mathbf{x}_i$ is an image, $y_i$ is the corresponding label, $\theta_S$ is the parameter set for the student network, and $\theta_{T}$ is the set for the teacher network. The loss $\mathcal{L}_{\text{sup}}$ is the alignment error between the network prediction and the annotation. For example in image classification task~\citep{mishra2017apprentice, shen2020amalgamating,polino2018model,cho2019efficacy}, it is normally a cross entropy loss. For object detection~\citep{liu2019learning, chen2017learning}, it includes bounding box regression as well. The distillation loss \(\mathcal{L}_{\text{distill}}\) quantifies how well the student mimics the pre-trained teacher, commonly implemented using KL divergence between softmax outputs~\citep{hinton2015distilling} or \(\ell_2\) distance between feature maps~\citep{romero2014fitnets}. While this approach demonstrates effectiveness with labeled data, its performance in unsupervised settings remains an open research question.

\subsection{Relational Representation Distillation}
\label{subsec:rrd}

Given an input image \( \mathbf{x}_i \), it is first mapped into features \( \mathbf{z}_i^T = f^T_\theta(\mathbf{x}_i) \) and \( \mathbf{z}_i^S = f^S_\theta(\mathbf{x}_i) \), where $\mathbf{z}_i^T, \mathbf{z}_i^S \in \mathbb{R}^d$ and $f^T_\theta, f^S_\theta$ denote the teacher and student networks, respectively. All features are $\ell_2$-normalized, \ie, $\mathbf{z}_i^T \leftarrow \frac{\mathbf{z}_i^T}{||\mathbf{z}_i^T||}$ and $\mathbf{z}_i^S \leftarrow \frac{\mathbf{z}_i^S}{||\mathbf{z}_i^S||}$, ensuring they lie on a unit hypersphere.

Let $\mathcal{M}=[\mathbf{m}_1, \ldots, \mathbf{m}_K]$ denote a memory bank where $K$ is the memory length and $\mathbf{m}_k\in \mathbb{R}^d$ is a feature vector. The memory $\mathcal{M}$ stores previous teacher features and is updated following a \textit{first-in-first-out} (FIFO) strategy: we add the teacher's features from the current batch while removing the oldest stored features per iteration. This buffer is critical for computing stable similarity distributions that capture relational structures between feature representations. Without stored references, similarity estimation would be limited to the current batch, restricting relational learning. While minimizing cross-entropy between student and teacher similarity distributions using $\mathcal{M}$ allows soft contrasting against random samples, direct teacher alignment is not enforced. To address this, we extend the memory bank to $\mathcal{M}^+=[\mathbf{m}_1, \ldots, \mathbf{m}_K, \mathbf{m}_{K+1}]$ by appending the teacher embedding $\mathbf{z}_i^T$ as $\mathbf{m}_{K+1}$. This ensures that the teacher's most recent representation is explicitly considered when computing similarity scores.

We define $\mathbf{p}^T(\mathbf{x}_i;\theta_T;\mathcal{M}^+)$ as the teacher similarity scores between the extracted teacher feature $\mathbf{z}_i^T$ and existing memory features $\mathbf{m}_j$ (for $j=1$ to $K+1$), represented as:

\begin{equation}
    \mathbf{p}^T(\mathbf{x}_i;\theta_T,\mathcal{M}^{+})=\left[p_1^T,\ldots,p^T_{K+1}\right]
\end{equation}

\noindent where

\begin{equation}
    {p_i^T} = \frac{\exp(\mathbf{z}_i^T \cdot \mathbf{m}_j/\tau_t)}{\sum\limits_{m \sim \mathcal{M}^{+}} \exp(\mathbf{z}_i^T \cdot \mathbf{m}/\tau_t)},
\end{equation}

\noindent and $(\cdot)$ denotes the \textit{inner product}, and $\tau_t$ is a temperature parameter for the teacher.
Similarly, we define $\mathbf{p}^S(\mathbf{x}_i;\theta_S,\mathcal{M}^+)$ as the student similarity scores between the extracted student feature $\mathbf{z}_i^S$ and existing memory features $\mathbf{m}_j$, represented as:

\begin{equation}
    \mathbf{p}^S(\mathbf{x}_i;\theta_S,\mathcal{M}^{+})=\left[p_1^S,\ldots,p^S_{K+1}\right]
\end{equation}

\noindent where

\begin{equation}
    {p_i^S} = \frac{\exp(\mathbf{z}_i^S \cdot \mathbf{m}_j/\tau_s)}{\sum\limits_{m \sim \mathcal{M}^{+}} \exp(\mathbf{z}_i^S \cdot \mathbf{m}/\tau_s)}
\end{equation}

\noindent and $\tau_s$ is a temperature parameter for the student.

Our distillation objective can be formulated as minimizing the KL divergence between the similarity scores of the teacher, $\mathbf{p}^T(\mathbf{x}_i;\theta_T, \mathcal{M}^+)$ and the student, $\mathbf{p}^{S}(\mathbf{x}_i;\theta_S,\mathcal{M}^+)$, over all the instances $\mathbf{x}_i$:

\begin{equation} 
\begin{aligned}  
    \hat{\theta}_{S}&=\argmin_{\theta_{S}} \sum_i^ND_{\text{KL}}(\mathbf{p}^T(\mathbf{x}_i;\theta_T, \mathcal{M}^+) \parallel \mathbf{p}^{S}(\mathbf{x}_i;\theta_S,\mathcal{M}^+)) 
    \\&=\argmin_{\theta_{S}} \sum_i^N H(\mathbf{p}^T(\mathbf{x}_i;\theta_T, \mathcal{M}^+), \mathbf{p}^{S}(\mathbf{x}_i;\theta_S,\mathcal{M}^+)) \\& + \redcancelto{\textcolor{red}{\text{constant}}}{H(\mathbf{p}^{T}(\mathbf{x}_i;\theta_T,\mathcal{M}^+))},
\end{aligned}
\end{equation}

\noindent where \( D_{\text{KL}} \) denotes the KL divergence between the teacher $\mathbf{p}^T(\mathbf{x}_i;\theta_T, \mathcal{M}^+)$ and the student $\mathbf{p}^{S}(\mathbf{x}_i;\theta_S,\mathcal{M}^+)$ distributions, \( H(\mathbf{p}^T(\mathbf{x}_i;\theta_T, \mathcal{M}^+), \mathbf{p}^{S}(\mathbf{x}_i;\theta_S,\mathcal{M}^+)) \) represents the cross-entropy between the teacher's and student's similarity distributions, and \( H(\mathbf{p}^T(\mathbf{x}_i;\theta_T, \mathcal{M}^+)) \) is the entropy of the teacher's similarity distribution. Since \( \mathbf{p}^{S}(\mathbf{x}_i;\theta_S,\mathcal{M}^+) \) will be used as a target, the gradient is clipped here, thus we only minimize the cross-entropy term \( H(\mathbf{p}^T(\mathbf{x}_i;\theta_T, \mathcal{M}^+), \mathbf{p}^{S}(\mathbf{x}_i;\theta_S,\mathcal{M}^+)) \):

\begin{equation}
	\hat{\theta}_{S} = \argmin_{\theta_{S}} \sum_{i=1}^{N} \mathcal{L}_{\textit{RRD}}(\mathbf{x}_i, \theta_{S}, \theta_{T}, \mathcal{M})
\end{equation}

\noindent where

\begin{equation} 
\begin{aligned}
	\mathcal{L}_{\textit{RRD}} 
	 = -\mathbf{p}^T(\mathbf{x}_i;\theta_T, \mathcal{M}^+) \cdot \log\mathbf{p}^{S}(\mathbf{x}_i;\theta_S,\mathcal{M}^+) \\
	\\= -\sum_{j=1}^{K+1} \frac{\exp(\mathbf{z}_i^T\cdot\mathbf{m}_j/\tau_t)}{\sum\limits_{k=1}^{K+1}\exp(\mathbf{z}_i^T\cdot \mathbf{m}_k/\tau_t)} \log\frac{\exp(\mathbf{z}_i^S\cdot\mathbf{m}_j/\tau_s)}{\sum\limits_{k=1}^{K+1}\exp(\mathbf{z}_i^S\cdot\mathbf{m}_k/\tau_s)}.
\end{aligned}
\end{equation}

\noindent Since we keep the teacher network frozen during training, teacher similarity scores $p_j^T$ directly influence corresponding student weights $p_j^S$. The $\ell_2$ normalization ensures similarity between $\mathbf{z}_i^T$ and $\mathbf{m}_{K+1}$ equals $\mathds{1}$ pre-softmax, making it dominate other $p_j^T$ values. This maximum weight for $p^S_{K+1}$ can be controlled via temperature $\tau_t$. The optimization aligns student features $\mathbf{z}_i^S$ with teacher features while maintaining contrast against memory features.

Note here that appending the current teacher embedding as $\mathbf{m}_{K+1}$ ensures a clear peak in $\mathbf{p}^T$, preventing the KL divergence from degenerating into weak contrastive alignment. As shown in our ablations (\Cref{sec:ablations}, \Cref{fig:ablation_memory}), performance improves with larger $K$ until plateauing around $K=16384$. Storing 128-d features requires only $\sim$600MB on ImageNet, allowing the bank to reside on GPU.

\subsubsection{Relation to \textit{InfoNCE} Loss}
As $\tau_t$ $\rightarrow 0$, the teacher's softmax distribution $\mathbf{p}^T$ becomes a one-hot vector with $p_{K + 1}^T=1$ and zeros elsewhere. This reduces our objective to:

\begin{equation}
   \begin{aligned}
    \mathcal{L}_\textit{{{NCE}}}
    =\sum_i^N-\log\frac{\exp(\mathbf{z}^T_i\cdot\mathbf{z}^S_i/\tau_s)}{\sum\limits_{\mathbf{m}\sim\mathcal{M}^+}\exp(\mathbf{z}^S_i\cdot\mathbf{m}/\tau_s)},
   \end{aligned}
\end{equation}

\noindent which matches the \textit{InfoNCE} loss~\citep{oord2019cpc}. This implements instance discrimination through $(K+1)$-way classification, separating different instances while enforcing identical representations for matching pairs.

\subsubsection{Relation to \textit{Kullback-Leibler} Divergence}
\citet{hinton2015distilling} defined the knowledge distillation loss via the \textit{Kullback–Leibler} (KL) divergence between the softened output distributions of the teacher and student networks:

\begin{equation}
\begin{aligned}
    \mathcal{L}_{\text{KL}} &= \sum\limits_{i=1}^{N} \tau^2 D_{\text{KL}}\Big(\sigma(y^T_i / \tau) \parallel \sigma(y^S_i / \tau) \Big) 
    \\&=\sum\limits_{i=1}^{N} \tau^2 \sum\limits_{c=1}^{C} \sigma\left(\frac{y^T_{i,c}}{\tau}\right) \log \frac{\sigma\left(\frac{y^T_{i,c}}{\tau}\right)} {\sigma\left(\frac{y^S_{i,c}}{\tau}\right)}
\end{aligned}
\label{eq:kldiv}
\end{equation}

\noindent where \( \sigma(x) \) denotes the softmax function, and \( y^T_i \), \( y^S_i \) represent the logits of the teacher and student networks, respectively, with \( y^S_{i,c} \) and \( y^T_{i,c} \) referring to their logit values for class \( c \), before applying the softmax function.
Both losses use KL divergence to align the teacher and student distributions. However, in Hinton's formulation the softmax is computed over \(C\) class logits (representing class predictions), while in our objective it is computed over \( (K+1)\) memory bank entries (representing similarity scores between features and memory bank entries).

\subsubsection{Full Objective}
Consistent with prior work in knowledge distillation, we formulate a full training objective that integrates supervised learning, standard KL divergence–based distillation, and our proposed loss. The full objective is given by:

\begin{equation}
\begin{aligned}
    \hat{\theta}_{S} &= \argmin_{\theta_{S}}\sum_{i}^{N} 
    \Big(
    \mathcal{L}_{\text{sup}}(\mathbf{x}_i, \theta_{S}, y_i)
    \\ &+ \lambda \cdot 
    \mathcal{L}_{{\text{KL}}}(\mathbf{x}_i, \theta_{S},  \theta_{T}) + \beta \cdot \mathcal{L}_{\textit{RRD}}(\mathbf{x}_i, \theta_{S}, \theta_{T}, \mathcal{M})
    \Big) 
\end{aligned}
\label{eq:full_loss}
\end{equation}

\noindent where \( \lambda \) and \( \beta \) balance the KL divergence and our proposed loss, respectively. The combination of losses provides complementary supervision: KD's soft targets provide direct class-level supervision through logit-space KL divergence, while our method ensures feature-space consistency. Unlike \citet{tian2022crd} which enforces strict instance-level discrimination that can push semantically similar samples too far apart, our method relaxes this constraint by focusing on preserving relative relationships in the feature space, allowing the student to maintain more nuanced similarities between instances while still learning discriminative representations.

\subsection{Information Bottleneck}
\label{subsec:info_bottleneck}

The dual-temperature formulation in our objective establishes an implicit \textit{information bottleneck} between the teacher and student similarity distributions (\cf \Cref{fig:info_bottleneck}). We set \( \tau_s > \tau_t \), where $\tau_t$ and $\tau_s$ denote the temperature parameters for the teacher and student, respectively. A smaller $\tau_t$ produces a \textit{sharper} teacher distribution $\mathbf{p}^T(\mathbf{x}_i;\theta_T,\mathcal{M}^+)$ that emphasizes dominant relational cues, while a larger $\tau_s$ yields a \textit{softer} student distribution $\mathbf{p}^S(\mathbf{x}_i;\theta_S,\mathcal{M}^+)$ that retains uncertainty over secondary similarities. This design enforces a selective transfer mechanism—only high-confidence, structurally salient relations from the teacher produce strong gradients in the student, while weaker or noisy relations are attenuated. Consequently, the student acts as a controlled filter, transmitting only the most informative relational cues from the teacher's representation space. This mechanism, which we refer to as the \textit{filtered information flow}, constrains the amount of information shared between teacher and student representations, forming a principled \textit{information bottleneck}. The effect of this temperature asymmetry and its empirical validation are analyzed in our ablation studies (\cf \Cref{sec:ablations}, \Cref{fig:ablation_temp}).

Formally, the mutual information between the teacher and student representations is bounded by the entropy difference between their respective similarity distributions:

\begin{equation}
    I(\mathbf{z}^T;\mathbf{z}^S) \leq H(\mathbf{p}^S(\mathbf{x}_i;\theta_S,\mathcal{M}^+)) - H(\mathbf{p}^T(\mathbf{x}_i;\theta_T,\mathcal{M}^+)),
\end{equation}

\noindent where \( H(\mathbf{p}^S(\mathbf{x}_i;\theta_S,\mathcal{M}^+)) > H(\mathbf{p}^T(\mathbf{x}_i;\theta_T,\mathcal{M}^+)) \) due to the softer student distribution. This entropy gap defines the effective information capacity of the distillation process, \ie, the larger the gap, the stronger the bottleneck, ensuring that our objective transmits only essential relational structure.

Intuitively, this bottleneck enforces a \textit{coarse-to-fine} transfer of representational knowledge: the teacher conveys sharp, high-fidelity relational signals, while the student absorbs them through smoother similarity distributions that generalize across related instances. This selective filtering prevents overfitting to instance-level details and encourages the student to preserve the underlying topology of the teacher's relational representation. Empirically, this \textit{filtered information flow} yields improved structural alignment (\Cref{tab:corralign}) and enhanced semantic organization (\Cref{tab:semantic}) compared to both conventional and contrastive distillation objectives.

\section{Experiments}
\label{sec:experiments}

We conduct extensive experiments across multiple benchmarks to evaluate the effectiveness and generality of our proposed objective\footnote{For clarity, we denote our method as ``\textit{RRD}'' when using only supervised learning (\ie, $\lambda=0$) and our relational loss, and ``\textit{RRD+KD}'' when combining all three loss components (\ie, $\lambda\neq0$).}. \Cref{subsec:setup} outlines the experimental setup, \Cref{subsec:results} presents quantitative results across benchmarks, and \Cref{subsec:representations} analyzes the learned representations and structural properties. Ablations are discussed in \Cref{sec:ablations}.

\subsection{Experimental Setup}
\label{subsec:setup}

We evaluate the proposed framework on both image classification and object detection tasks using five standard benchmarks: CIFAR-100~\citep{krizhevsky2009cifar}, ImageNet ILSVRC-2012~\citep{deng2009imagenet}, STL-10~\citep{coates2011importance}, Tiny ImageNet~\citep{deng2009imagenet}, and MS-COCO~\citep{li2015coco}. Following prior work~\citep{tian2022crd}, we evaluate thirteen teacher–student architecture pairs with varying capacity gaps to assess generalization across different model families. For classification, we adopt the implementation protocol of~\citep{tian2022crd}, while for detection we follow~\citep{zhao2022dkd,chen2021reviewkd}.
To ensure dimensional consistency and preserve relational information during feature alignment, both teacher and student features are passed through \textit{projection heads}—two-layer MLPs (512 hidden, 128 output)—that nonlinearly project intermediate representations into a shared embedding space for computing relational similarity distributions, improving KL alignment stability and structural coherence as observed in prior work~\citep{tian2022crd,miles2024understanding}.
The memory bank size is fixed at $K = 16{,}384$. The temperature parameters are set to $\tau_t = 0.02$ for the teacher and $\tau_s = 0.1$ for the student, reflecting the asymmetric softening central to our \textit{information bottleneck} formulation (\Cref{fig:info_bottleneck} validated in \Cref{fig:ablation_temp}). The distillation weight $\lambda$ is fixed at $0.9$ for CIFAR-100 and $1.0$ for ImageNet, while $\beta$ is set to $1.5$ and $1.0$ respectively. When combined with standard KD, the temperature for the logit-space KL divergence is $\tau = 4$.

\subsection{Main Results}
\label{subsec:results}

We first evaluate RRD on image classification (CIFAR-100, ImageNet) and object detection (MS-COCO). Across all tasks, our objective consistently improves over baseline distillation methods and shows strong complementarity when combined with Hinton's KD \citep{hinton2015distilling}, confirming its ability to capture transferable relational structure.

\subsubsection{Results on CIFAR-100}

\begin{table*}[!t]
\centering
\caption{\textbf{Classification distillation results on CIFAR-100.} Test top-1 accuracy (\%) for different teacher–student pairs, grouped into same-architecture and different-architecture setups. The best and second-best results (per column) are \textbf{bolded} and \underline{underlined}. Results for baselines are from \citet{tian2022crd}; ours are averaged over five runs. \textbf{Architecture abbreviations}: W: WideResNet, rn: resnet, RN: ResNet, MN: MobileNet, SN: ShuffleNet.}
\label{tab:cifar100}
\setlength{\tabcolsep}{0.3mm}
\resizebox{\textwidth}{!}{%
\begin{tabular}{lccccccccccccc}
\toprule
& \multicolumn{7}{c}{\textbf{Same architecture}} & \multicolumn{6}{c}{\textbf{Different architecture}} \\
\cmidrule(lr){2-8} \cmidrule(lr){9-14}
Teacher & WRN-40-2 & WRN-40-2 & rn-56 & rn-110 & rn-110 & rn-32x4 & VGG-13 & VGG-13 & RN-50 & RN-50 & RN-32x4 & RN-32x4 & WRN-40-2 \\
Student & WRN-16-2 & WRN-40-1 & rn-20 & rn-20 & rn-32 & rn-8x4 & VGG-8 & MN-v2 & MN-v2 & VGG-8 & SN-v1 & SN-v2 & SN-v1 \\
\midrule
\textit{Teacher} & 75.61 & 75.61 & 72.34 & 74.31 & 74.31 & 79.42 & 74.64 & 74.64 & 79.34 & 79.34 & 79.42 & 79.42 & 75.61 \\
\textit{Student} & 73.26 & 71.98 & 69.06 & 69.06 & 71.14 & 72.50 & 70.36 & 64.60 & 64.60 & 70.36 & 70.50 & 71.82 & 70.50 \\
KD \citep{hinton2015distilling} & 74.92 & 73.54 & 70.66 & 70.67 & 73.08 & 73.33 & 72.98 & 67.37 & 67.35 & 73.81 & 74.07 & 74.45 & 74.83 \\
FitNet \citep{romero2014fitnets} & 73.58 & 72.24 & 69.21 & 68.99 & 71.06 & 73.50 & 71.02 & 64.14 & 63.16 & 70.69 & 73.59 & 73.54 & 73.73 \\
AT \citep{zagoruyko2016paying} & 74.08 & 72.77 & 70.55 & 70.22 & 72.31 & 73.44 & 71.43 & 59.40 & 58.58 & 71.84 & 71.73 & 72.73 & 73.32 \\
SP \citep{tung2019similarity} & 73.83 & 72.43 & 69.67 & 70.04 & 72.69 & 72.94 & 72.68 & 66.30 & 68.08 & 73.34 & 73.48 & 74.56 & 74.52 \\
CC \citep{peng2019correlation} & 73.56 & 72.21 & 69.63 & 69.48 & 71.48 & 72.97 & 70.81 & 64.86 & 65.43 & 70.25 & 71.14 & 71.29 & 71.38 \\
RKD \citep{park2019relational} & 73.35 & 72.22 & 69.61 & 69.25 & 71.82 & 71.90 & 71.48 & 64.52 & 64.43 & 71.50 & 72.28 & 73.21 & 72.21 \\
FSP \citep{yim2017gift} & 72.91 & n/a & 69.95 & 70.11 & 71.89 & 72.62 & 70.33 & 58.16 & 64.96 & 71.28 & 74.12 & 74.68 & 76.09 \\
OFD \citep{heo2019ofd} & 75.24 & 74.33 & 70.38 & n/a & 73.23 & 74.95 & 73.95 & 69.48 & 69.04 & n/a & \underline{75.98} & \underline{76.82} & 75.85 \\
CRD \citep{tian2022crd} & 75.48 & 74.14 & 71.16 & 71.46 & 73.48 & \underline{75.51} & 73.94 & 69.73 & 69.11 & 74.30 & 75.11 & 75.65 & 76.05 \\
CRD+KD \citep{tian2022crd} & 75.64 & 74.38 & 71.63 & 71.56 & \underline{73.75} & 75.46 & \underline{74.29} & \underline{69.94} & 69.54 & \textbf{74.97} & 75.12 & 76.05 & \underline{76.27} \\
RRD (ours) & \cellcolor{pearDarker!08}\textbf{75.85} & \cellcolor{pearDarker!08}\underline{74.61} & \cellcolor{pearDarker!08}\underline{71.89} & \cellcolor{pearDarker!08}\textbf{71.92} & \cellcolor{pearDarker!08}73.73 & \cellcolor{pearDarker!08}\textbf{75.77} & \cellcolor{pearDarker!08}74.01 & \cellcolor{pearDarker!08}69.61 & \cellcolor{pearDarker!08}\textbf{70.11} & \cellcolor{pearDarker!08}74.30 & \cellcolor{pearDarker!08}75.60 & \cellcolor{pearDarker!08}76.31 & \cellcolor{pearDarker!08}75.98 \\
RRD+KD (ours) & \cellcolor{pearDarker!08}\underline{75.67} & \cellcolor{pearDarker!08}\textbf{74.68} & \cellcolor{pearDarker!08}\textbf{72.03} & \cellcolor{pearDarker!08}\underline{71.75} & \cellcolor{pearDarker!08}\textbf{73.96} & \cellcolor{pearDarker!08}75.53 & \cellcolor{pearDarker!08}\textbf{74.37} & \cellcolor{pearDarker!08}\textbf{69.99} & \cellcolor{pearDarker!08}\underline{69.65} & \cellcolor{pearDarker!08}\underline{74.53} & \cellcolor{pearDarker!08}\textbf{76.68} & \cellcolor{pearDarker!08}\textbf{76.87} & \cellcolor{pearDarker!08}\textbf{76.64} \\
\bottomrule
\end{tabular}%
}
\end{table*}

\Cref{tab:cifar100} reports top-1 accuracy for same-architecture and cross-architecture teacher-student pairs. RRD consistently outperforms classical KD \citep{hinton2015distilling} and contrastive CRD \citep{tian2022crd}, showing that preserving relative similarity distributions guides the student more effectively than strict instance discrimination. Adding Hinton's KD \citep{hinton2015distilling} typically yields further gains, since KD supplies class-level supervision through logit-space KL divergence while RRD enforces relational consistency in feature space. On a few close-capacity or same-architecture pairs the combined variant slightly trails standalone RRD, a known effect also reported for \citet{tian2022crd} that arises when the KD term contributes conflicting rather than complementary gradients.

\subsubsection{Results on ImageNet}

\begin{table*}[!t]
\centering
\setlength{\tabcolsep}{0.8mm}
\caption{\textbf{Classification distillation results on ImageNet ILSVRC-2012.} Student top-1 accuracy (\%) on the ImageNet validation set under various teacher–student pairs. Results are from \citet{tian2022crd}; ours are based on a single run.}
\begin{tabular}{lcccccccccc}
\toprule
& \textit{Teacher} & \textit{Student} & KD \cite{hinton2015distilling} & AT \cite{zagoruyko2016paying} & SP \cite{tung2019similarity} & CC \cite{peng2019correlation} & RKD \cite{park2019relational} & CRD \cite{tian2022crd} & RRD & RRD+KD \\
\midrule
ResNet-34$\rightarrow$ResNet-18 & 73.31 & 69.75  & 70.67 & 71.03 & 70.62 & 69.96 & 70.40 & 71.17 & \cellcolor{pearDarker!08}\textbf{72.03} & \cellcolor{pearDarker!08}\underline{71.99}  \\
ResNet-50$\rightarrow$ResNet-18 & 76.16 & 69.75  & 71.29 & 71.18 & 71.08 & n/a & n/a & 71.25 & \cellcolor{pearDarker!08}\textbf{71.97} & \cellcolor{pearDarker!08}\underline{71.88}  \\
ResNet-50$\rightarrow$MobileNet-v2 & 76.16 & 69.63  & 70.49 & 70.18 & n/a & n/a & 68.50 & 69.07 & \cellcolor{pearDarker!08}\underline{71.54} & \cellcolor{pearDarker!08}\textbf{71.56}  \\
\bottomrule
\end{tabular}
\label{tab:imagenet}
\end{table*}

\Cref{tab:imagenet} shows that RRD scales to large benchmarks such as ImageNet, sustaining strong and consistent accuracy across diverse architectures. The advantage holds for both same- and cross-architecture pairs, and RRD surpasses KD \citep{hinton2015distilling}, CRD \citep{tian2022crd}, and their combination in every evaluated setting, even when applied on its own, underscoring its robustness and broad applicability.

\subsubsection{Results on COCO}

\begin{table*}[!t]
\centering
\caption{\textbf{Object detection distillation results on COCO.} Detection performance (AP, AP$_{50}$, AP$_{75}$) of student detectors trained with different distillation methods using Faster R-CNN on COCO \texttt{val2017}. Results are from \citet{zhao2022dkd}; ours are based on a single run. \textbf{Architecture abbreviations}: RN: ResNet, MN: MobileNet.}
\label{tab:coco}
\setlength{\tabcolsep}{1.85mm}
\begin{tabular}{lcccccccccc}
\toprule
\multirow{2}{*}{Method} & \multicolumn{3}{c}{RN-101 $\rightarrow$ RN-18} & \multicolumn{3}{c}{RN-101 $\rightarrow$ RN-50} & \multicolumn{3}{c}{RN-50 $\rightarrow$ MN-v2} \\
\cmidrule(lr){2-4} \cmidrule(lr){5-7} \cmidrule(lr){8-10}
& AP & AP$_{50}$ & AP$_{75}$ & AP & AP$_{50}$ & AP$_{75}$ & AP & AP$_{50}$ & AP$_{75}$ \\
\midrule
\textit{Teacher} & 42.04 & 62.48 & 45.88 & 42.04 & 62.48 & 45.88 & 40.22 & 61.02 & 43.81 \\
\textit{Student} & 33.26 & 53.61 & 35.26 & 37.93 & 58.84 & 41.05 & 29.47 & 48.87 & 30.90 \\
KD \citep{hinton2015distilling} & 33.97 & 54.66 & 36.62 & 38.35 & 59.41 & 41.71 & 30.13 & 50.28 & 31.35 \\
FitNet \citep{romero2014fitnets} & 34.13 & 54.16 & 36.71 & 38.76 & 59.62 & 41.80 & 30.20 & 49.80 & 31.69 \\
ReviewKD \citep{chen2021reviewkd} & 36.75 & 56.72 & 34.00 & \textbf{40.36} & 60.97 & \textbf{44.08} & 33.71 & 53.15 & 36.13 \\
DKD \citep{zhao2022dkd} & 35.05 & 56.60 & 37.54 & 39.25 & 60.90 & 42.73 & 32.34 & 53.77 & 34.01 \\
RRD (ours) & \cellcolor{pearDarker!08}\textbf{36.85} & \cellcolor{pearDarker!08}\textbf{57.10} & \cellcolor{pearDarker!08}\textbf{39.20} & \cellcolor{pearDarker!08}40.15 & \cellcolor{pearDarker!08}\textbf{61.00} & \cellcolor{pearDarker!08}43.90 & \cellcolor{pearDarker!08}\textbf{33.90} & \cellcolor{pearDarker!08}\textbf{54.20} & \cellcolor{pearDarker!08}\textbf{36.00} \\
\bottomrule
\end{tabular}
\end{table*}

\Cref{tab:coco} extends the evaluation to object detection with Faster R-CNN models trained under different distillation methods on MS-COCO \citep{li2015coco}. By transferring relational cues from teacher feature maps, RRD improves detection accuracy across teacher–student pairs while remaining stable during fine-tuning. Its results are competitive with or better than advanced methods such as ReviewKD \citep{chen2021reviewkd} and DKD \citep{zhao2022dkd}, indicating that relational structure also benefits spatially localized prediction.

\subsubsection{Efficiency}

\begin{table*}[!t]
\centering
\setlength{\tabcolsep}{0.8mm}
\caption{\textbf{Computational cost on CIFAR-100.} Comparison of training latency (ms) and parameter counts for different distillation frameworks. We use a ResNet-32x4 teacher and ResNet-8x4 student within the \texttt{MDistiller} framework \citep{zhao2022dkd} on an NVIDIA RTX 6000 GPU.}
\label{tab:efficiency}
\begin{tabular}{lccccccccc}
\toprule
& KD \citep{hinton2015distilling} & FitNet \citep{romero2014fitnets} & AT \citep{zagoruyko2016paying} & RKD \citep{park2019relational} & CRD \citep{tian2022crd} & OFD \citep{heo2019ofd} & ReviewKD \citep{chen2021reviewkd} & DKD \citep{zhao2022dkd} & RRD (ours) \\
\midrule
Time/batch & 8 & 8 & 10 & 17 & 19 & 20 & 14 & 9 & \cellcolor{pearDarker!08}15 \\
Params & 0 & 16,768 & 0 & 0 & 12,865,797 & 86,912 & 1,808,774 & 0 & \cellcolor{pearDarker!08}2,162,944 \\
\bottomrule
\end{tabular}
\end{table*}

\Cref{tab:efficiency} reports the training cost, including latency and parameter overhead, of different distillation methods on CIFAR-100 using a ResNet-32x4 teacher and ResNet-8x4 student. RRD runs at 15ms per batch with 2.16M additional parameters. It is slower than parameter-free logit methods such as KD \citep{ma2022distilling} and DKD \citep{zhao2022dkd} but competitive with feature-based approaches: CRD \cite{tian2022crd} needs 12.87M parameters at 19ms per batch, ReviewKD \citep{chen2021reviewkd} runs at 14ms with 1.81M parameters, and OFD \citep{heo2019ofd} adds only 0.087M parameters yet is the slowest at 20ms.

\subsection{Representation Analysis}
\label{subsec:representations}

We next analyze the learned representations to assess their structural and semantic fidelity. 

\subsubsection{Visualization of Inter-class Correlations}

\begin{figure*}[!t]
    \centering
    \begin{subfigure}[t]{0.15\textwidth}
        \centering
        \includegraphics[width=\linewidth]{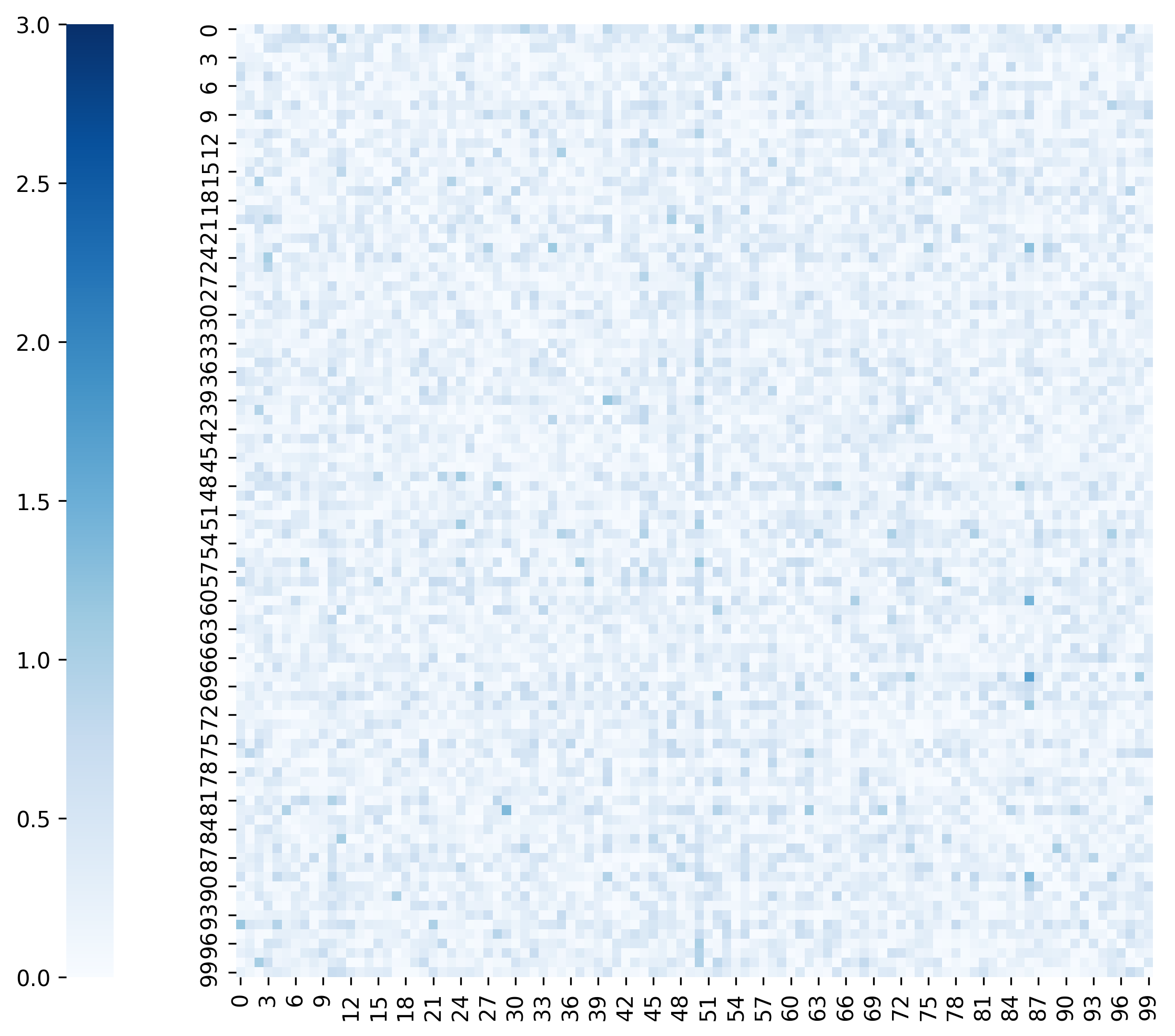}
        \caption{Vanilla}
        \label{fig:corr_main_vanilla}
    \end{subfigure}
    \hfill
    \begin{subfigure}[t]{0.15\textwidth}
        \centering
        \includegraphics[width=\linewidth]{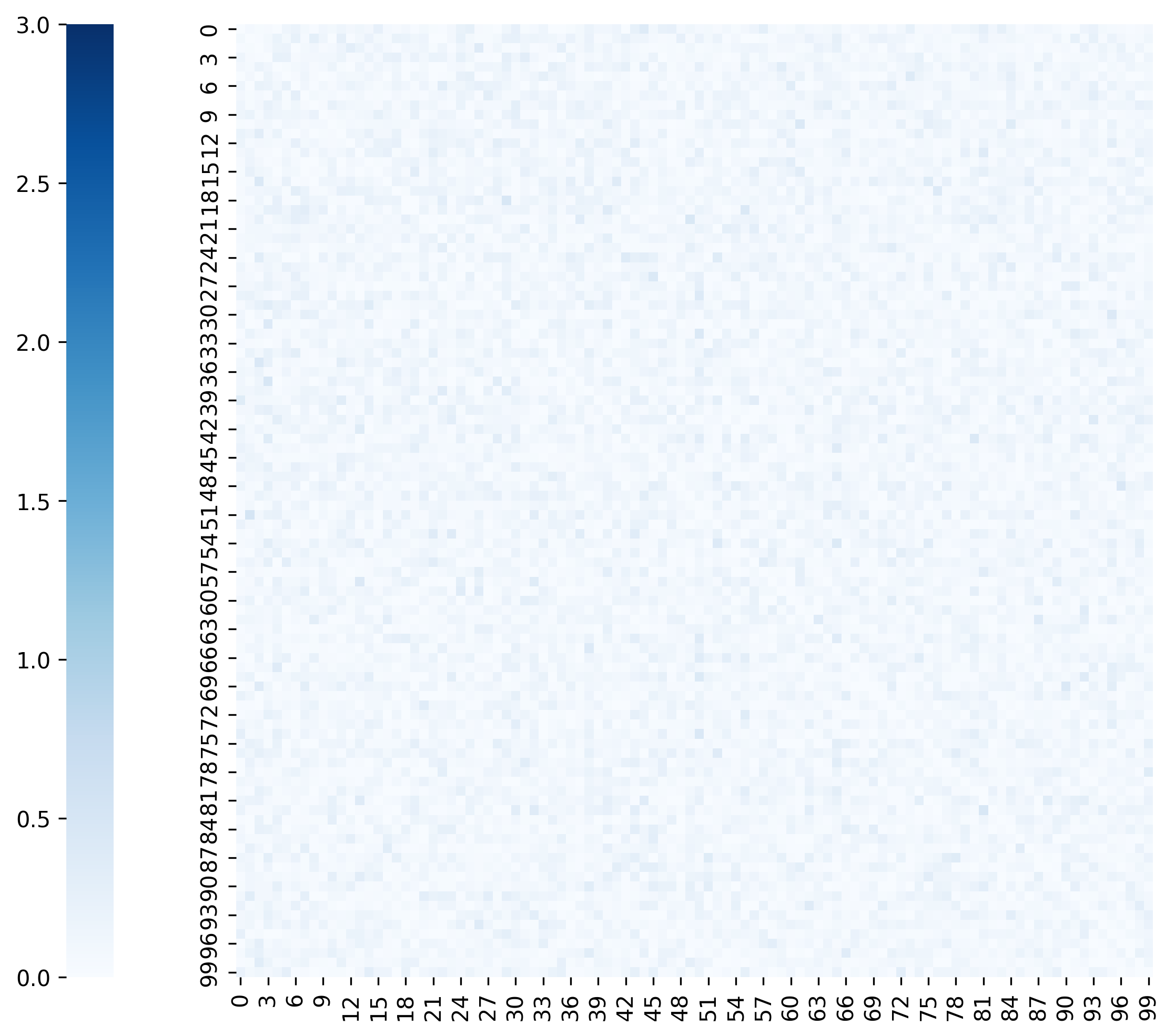}
        \caption{KD}
        \label{fig:corr_main_kd}
    \end{subfigure}
    \hfill
    \begin{subfigure}[t]{0.15\textwidth}
        \centering
        \includegraphics[width=\linewidth]{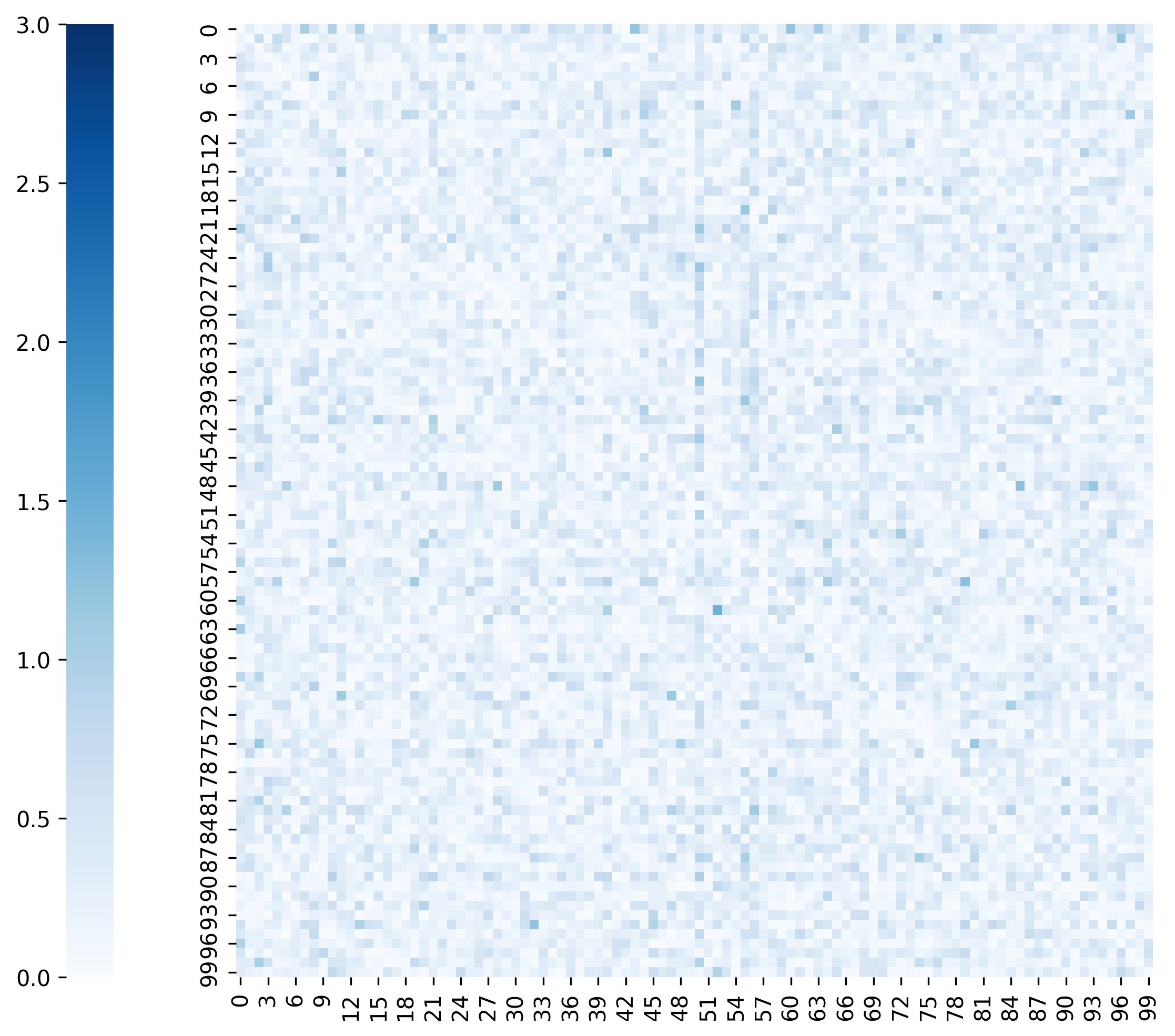}
        \caption{CRD}
        \label{fig:corr_main_crd}
    \end{subfigure}
    \hfill
    \begin{subfigure}[t]{0.15\textwidth}
        \centering
        \includegraphics[width=\linewidth]{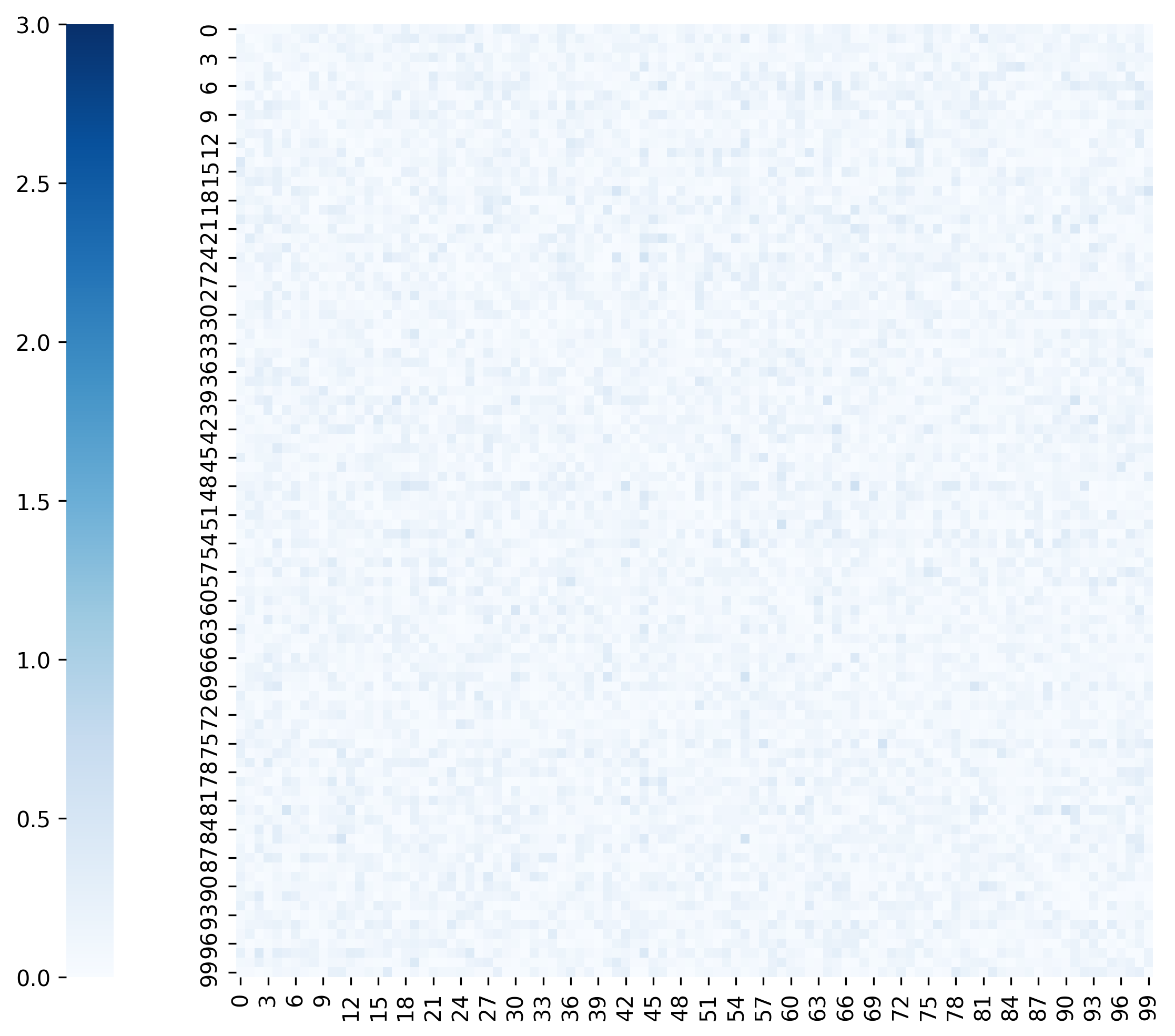}
        \caption{CRD+KD}
        \label{fig:corr_main_crd_kd}
    \end{subfigure}
    \hfill
    \begin{subfigure}[t]{0.15\textwidth}
        \centering
        \includegraphics[width=\linewidth]{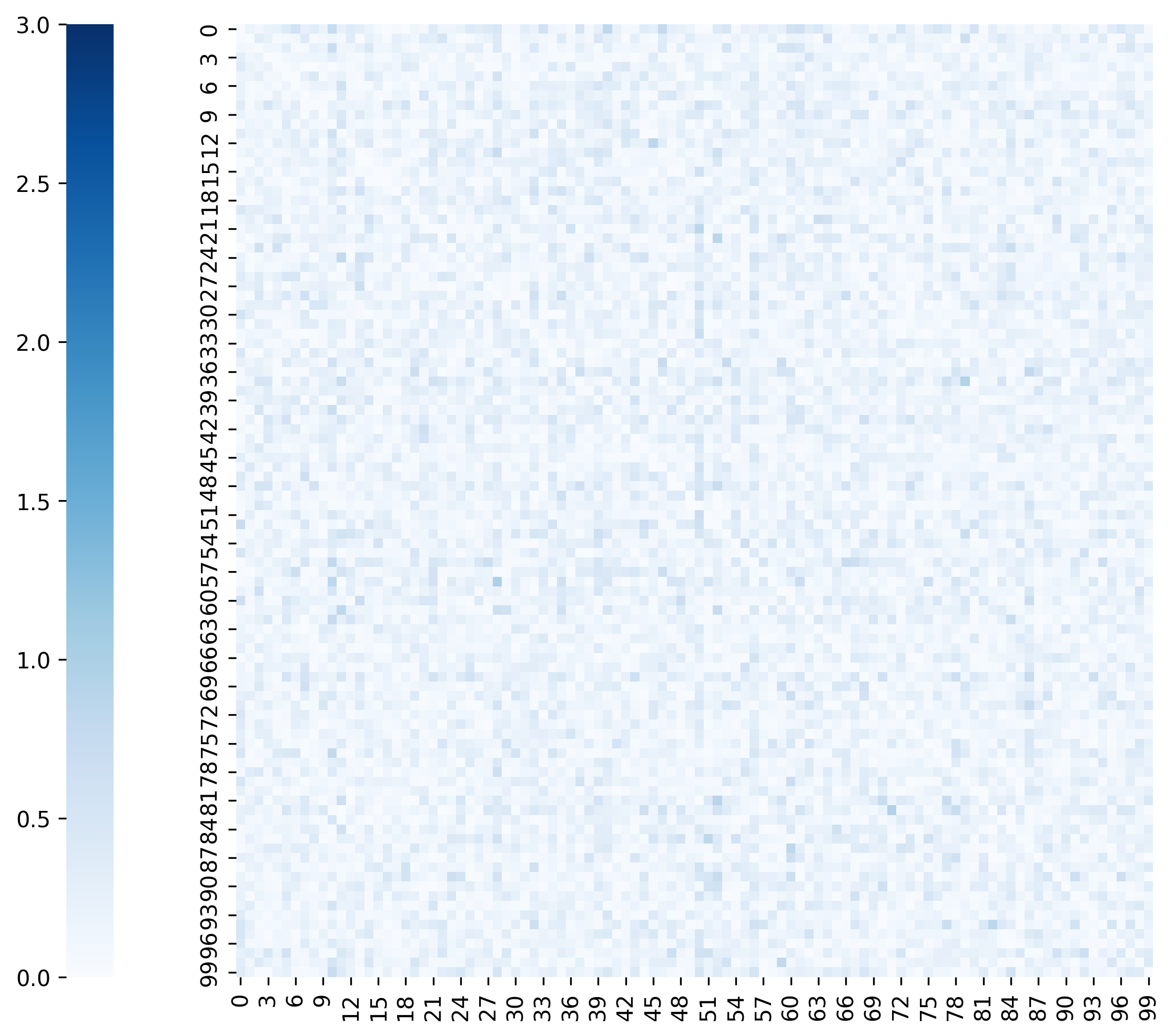}
        \caption{RRD}
        \label{fig:corr_main_rrd}
    \end{subfigure}
    \hfill
    \begin{subfigure}[t]{0.15\textwidth}
        \centering
        \includegraphics[width=\linewidth]{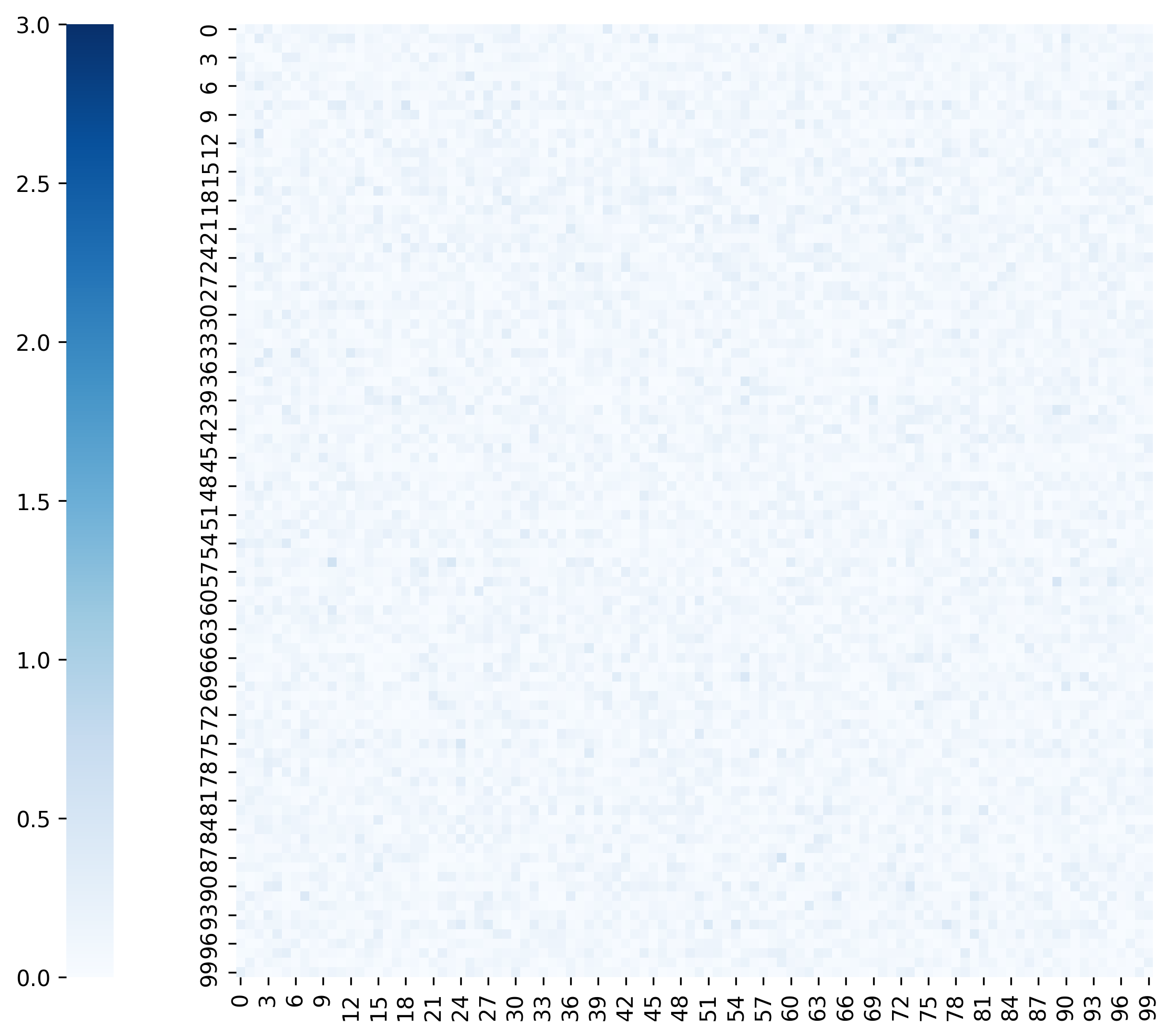}
        \caption{RRD+KD}
        \label{fig:corr_main_rrd_kd}
    \end{subfigure}
    \caption{\textbf{Correlation alignment on CIFAR-100.}
    \subref{fig:corr_main_vanilla} Vanilla (Mean: 0.24, Max: 1.66),
    \subref{fig:corr_main_kd} KD \citep{hinton2015distilling} (Mean: 0.09, Max: 0.49),
    \subref{fig:corr_main_crd} CRD \citep{tian2022crd} (Mean: 0.23, Max: 1.56),
    \subref{fig:corr_main_crd_kd} CRD+KD \citep{tian2022crd} (Mean: 0.10, Max: 0.57),
    \subref{fig:corr_main_rrd} RRD (ours) (Mean: 0.18, Max: 0.99),
    \subref{fig:corr_main_rrd_kd} RRD+KD (ours) (Mean: 0.07, Max: 0.55).
    Correlation matrix comparison between teacher (WRN-40-2) and student (WRN-40-1) logits. Lower values indicate stronger alignment of inter-class relations.}
    \label{fig:correlations}
\end{figure*}

\Cref{fig:correlations} compares the correlation-matrix differences between teacher and student logits. RRD aligns correlation structure more closely than models trained without distillation or with alternative methods \citep{hinton2015distilling,tian2022crd}. On its own it surpasses CRD \citep{tian2022crd}, indicating stronger structural preservation, and combining it with Hinton's KD \citep{hinton2015distilling} tightens the alignment further. This coherence suggests that RRD leads the student to internalize the teacher's feature geometry rather than memorizing isolated logits.




\subsubsection{Transferability of Representations}

\begin{table*}[!t]
\centering
\caption{\textbf{Transfer learning performance results on STL-10 and Tiny ImageNet.} Top-1 accuracy (\%) of a WRN-16-2 student distilled from WRN-40-2 (CIFAR-100). Results for baselines are from \citet{tian2022crd}; ours are averaged over five runs.}
\label{tab:transferability}
\setlength{\tabcolsep}{1.6mm}
\begin{tabular}{lccccccccc}
\toprule
& \textit{Teacher} & \textit{Student} & KD \citep{hinton2015distilling} & AT \citep{zagoruyko2016paying} & FitNet \citep{romero2014fitnets} & CRD \citep{tian2022crd} & CRD+KD \citep{tian2022crd} & RRD & RRD+KD \\
\midrule
STL-10 & 68.6 & 69.7 & 70.9 & 70.7 & 70.3 & 71.6 & \textbf{72.2} & \cellcolor{pearDarker!08}\underline{72.0} & \cellcolor{pearDarker!08}\underline{72.0} \\
Tint ImageNet & 31.5 & 33.7 & 33.9 & 34.2 & 33.5 & \textbf{35.6} & \underline{35.5} & \cellcolor{pearDarker!08}\underline{35.5} & \cellcolor{pearDarker!08}35.2 \\
\bottomrule
\end{tabular}
\end{table*}

\Cref{tab:transferability} evaluates a WRN-16-2 student distilled from a WRN-40-2 teacher used as a frozen feature extractor. RRD delivers strong and consistent transfer to unseen datasets, showing that its representations encode relational semantics rather than overfitting to task-specific decision boundaries or local class structure.

\subsubsection{Correlation Matrix Alignment}

\begin{table}[!t]
\centering
\caption{\textbf{Correlation matrix alignment metrics on CIFAR-100.} Quantitative comparison of relational alignment between teacher (WRN-40-2) and student (WRN-40-1) embeddings using Frobenius distance, Pearson correlation, and SSIM.}
\label{tab:corralign}
\setlength{\tabcolsep}{1.6mm}
\begin{tabular}{lccc}
\toprule
Method & Frobenius $\downarrow$ & Pearson $\uparrow$ & SSIM $\uparrow$ \\
\midrule
Vanilla & 10.491 & 0.951 & 0.935 \\
KD~\citep{hinton2015distilling} & \underline{2.868} & \underline{0.994} & \underline{0.992} \\
CRD~\citep{tian2022crd} & 8.862 & 0.946 & 0.938 \\
CRD+KD~\citep{tian2022crd} & 3.265 & 0.992 & 0.989 \\
RRD (ours) & \cellcolor{pearDarker!08}6.969 & \cellcolor{pearDarker!08}0.970 & \cellcolor{pearDarker!08}0.964 \\
RRD+KD (ours) & \cellcolor{pearDarker!08}\textbf{2.720} & \cellcolor{pearDarker!08}\textbf{0.995} & \cellcolor{pearDarker!08}\textbf{0.993} \\
\bottomrule
\end{tabular}
\end{table}
\begin{table}[!t]
\centering
\caption{\textbf{Semantic similarity preservation on CIFAR-100.} Evaluation of intra-/inter-class structure, NMI, and retrieval mAP@5 of student embeddings under different distillation methods.}
\setlength{\tabcolsep}{1.4mm}
\label{tab:semantic}
\begin{tabular}{lcccc}
\toprule
Method & Intra $\downarrow$ & Inter $\uparrow$ & NMI $\uparrow$ & mAP@5 $\uparrow$ \\
\midrule
Vanilla & 0.945 & 0.866 & 0.551 & 81.0 \\
KD~\citep{hinton2015distilling} & \underline{0.914} & \underline{1.012} & \underline{0.546} & \underline{85.4} \\
CRD~\citep{tian2022crd} & 0.910 & 0.915 & 0.509 & 84.6 \\
CRD+KD~\citep{tian2022crd} & 0.973 & 0.926 & 0.463 & 85.2 \\
RRD (ours) & \cellcolor{pearDarker!08}0.928 & \cellcolor{pearDarker!08}0.909 & \cellcolor{pearDarker!08}0.544 & \cellcolor{pearDarker!08}83.9 \\
RRD+KD (ours) & \cellcolor{pearDarker!08}\textbf{0.902} & \cellcolor{pearDarker!08}\textbf{1.023} & \cellcolor{pearDarker!08}\textbf{0.562} & \cellcolor{pearDarker!08}\textbf{85.9} \\
\bottomrule
\end{tabular}
\end{table}
\begin{figure*}[!t]
    \centering
    \begin{subfigure}[t]{0.22\textwidth}
        \centering
        \includegraphics[width=\linewidth]{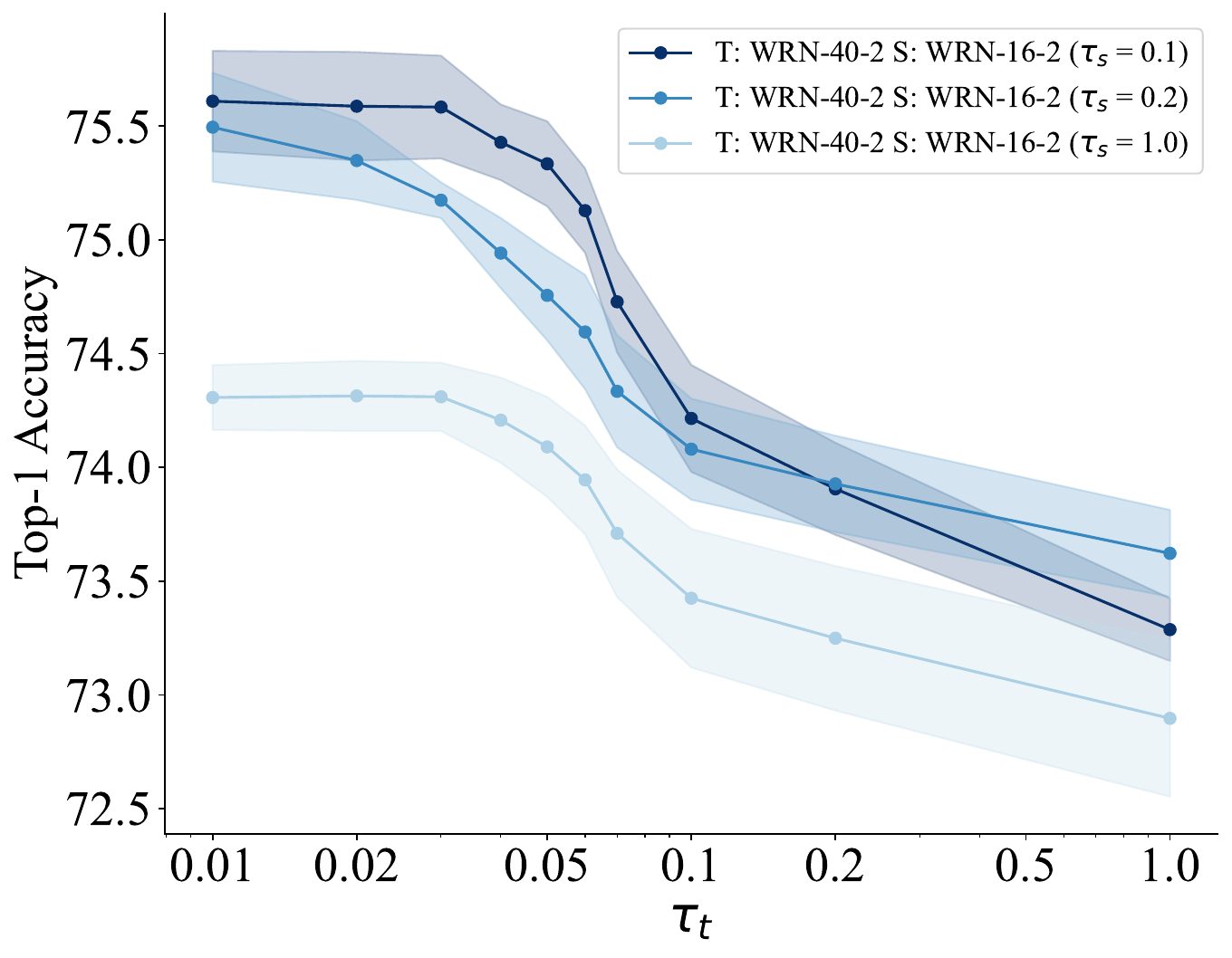}
        \caption{Temperatures $(\tau_t, \tau_s)$}
        \label{fig:ablation_temp}
    \end{subfigure}
    \hfill
    \begin{subfigure}[t]{0.22\textwidth}
        \centering
        \includegraphics[width=\linewidth]{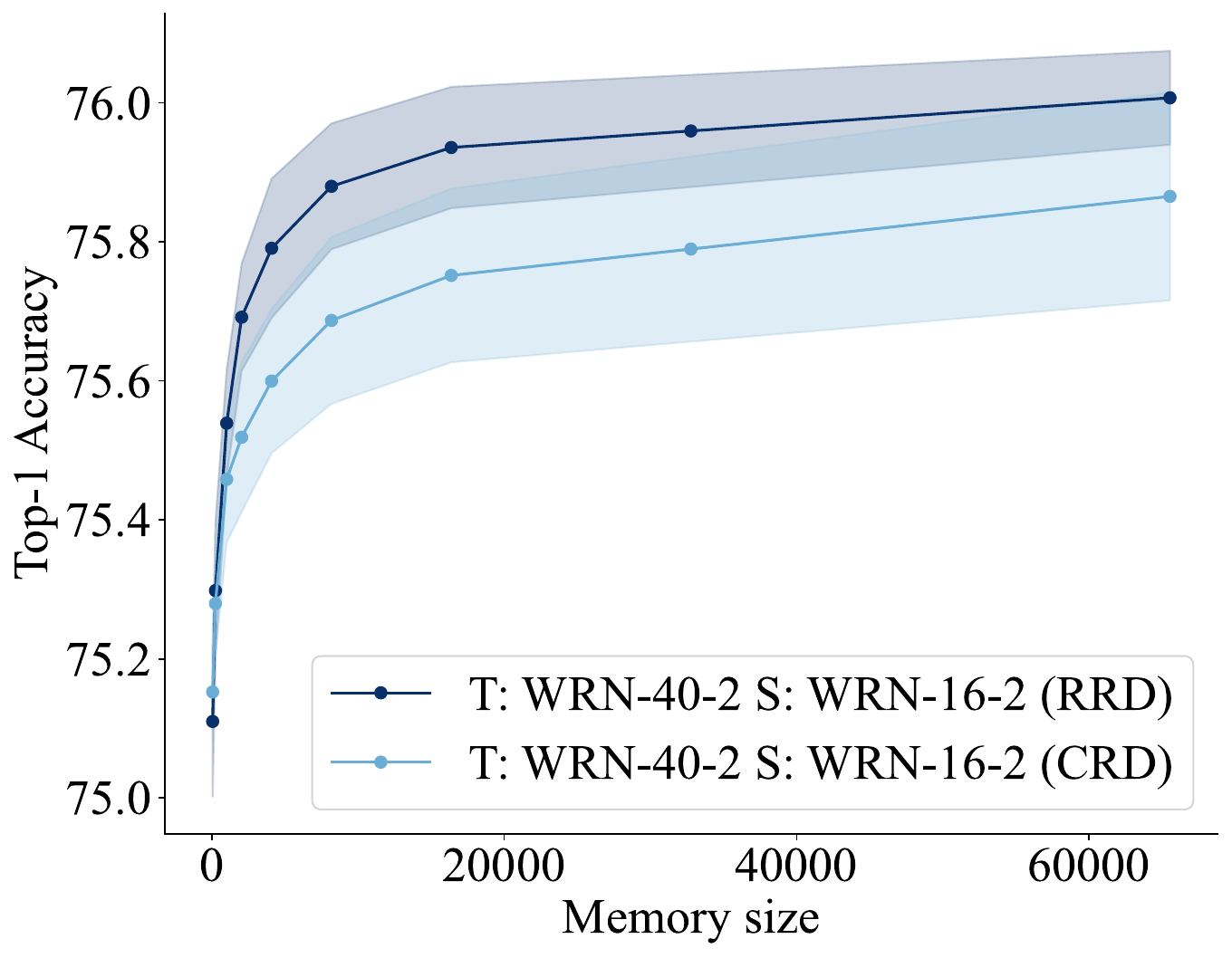}
        \caption{Memory $K$}
        \label{fig:ablation_memory}
    \end{subfigure}
    \hfill
    \begin{subfigure}[t]{0.22\textwidth}
        \centering
        \includegraphics[width=\linewidth]{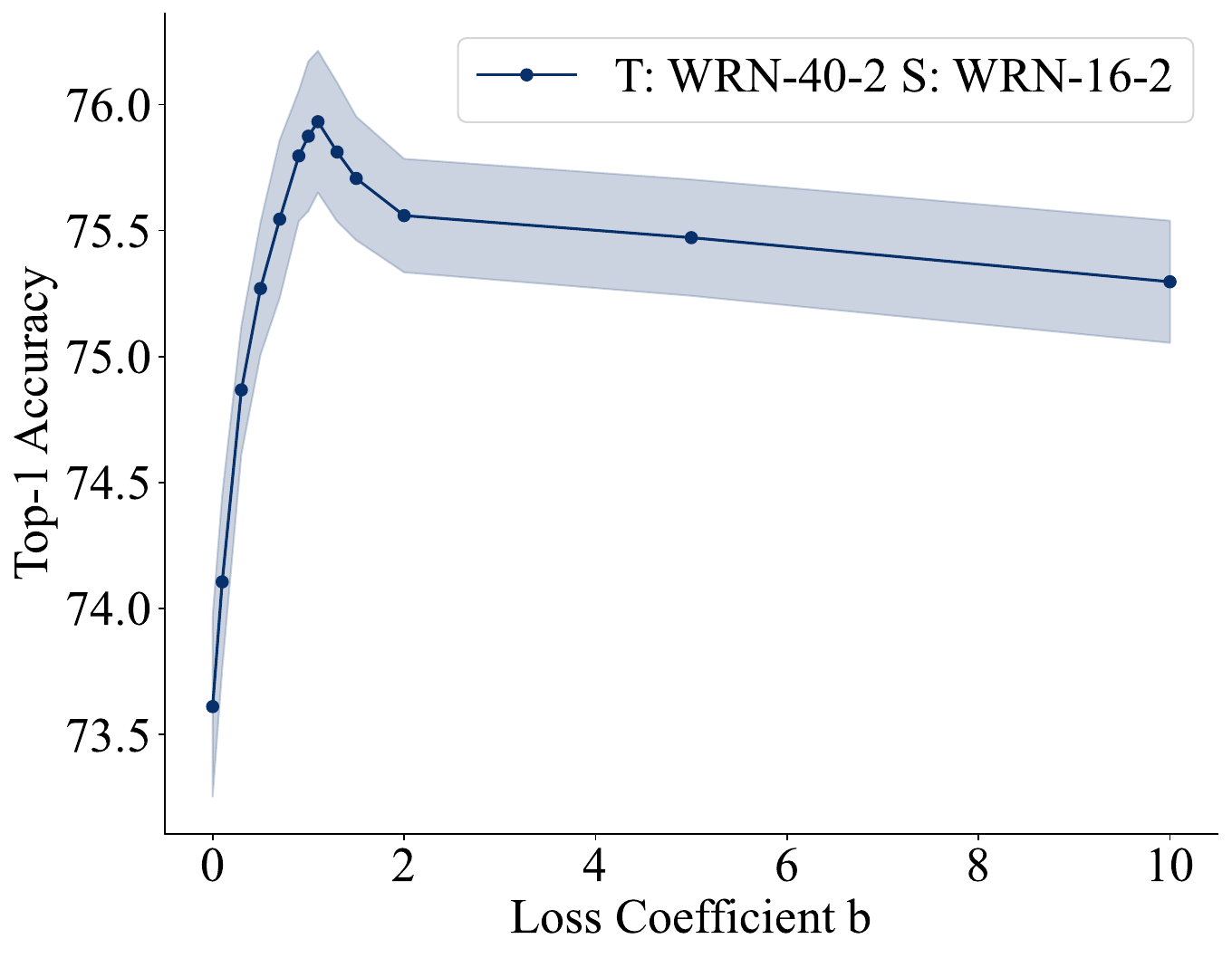}
        \caption{Loss coefficient $\beta$}
        \label{fig:ablation_beta}
    \end{subfigure}
    \hfill
    \begin{subfigure}[t]{0.22\textwidth}
        \centering
        \includegraphics[width=\linewidth]{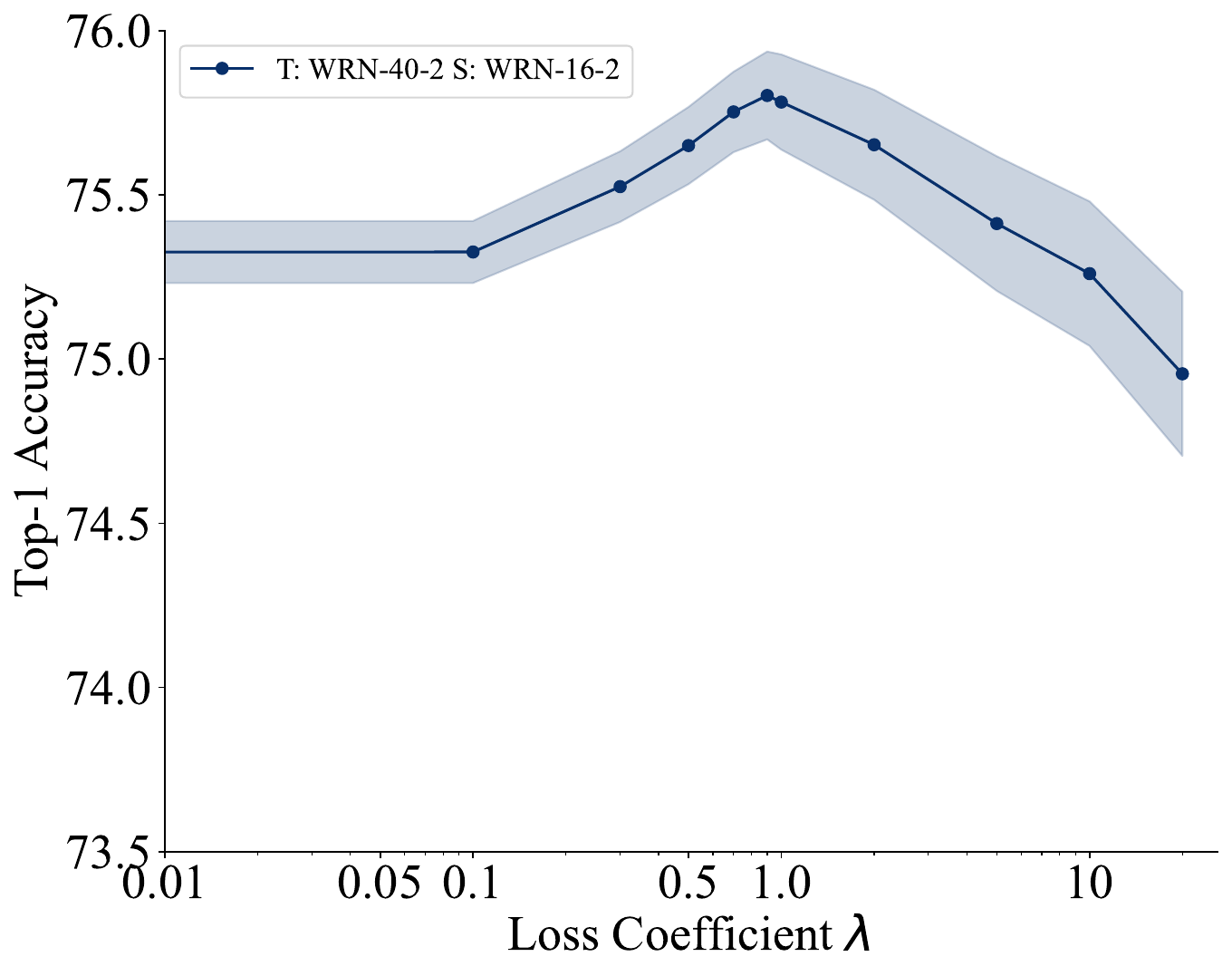}
        \caption{Loss coefficient $\lambda$}
        \label{fig:ablation_lambda}
    \end{subfigure}
    \caption{\textbf{Ablation study.}
    \subref{fig:ablation_temp} Temperature parameters $(\tau_t, \tau_s)$.
    \subref{fig:ablation_memory} Memory size $(K)$.
    \subref{fig:ablation_beta} Loss coefficient $\beta$.
    \subref{fig:ablation_lambda} Loss coefficient $\lambda$.
    }
    \label{fig:ablations}
\end{figure*}

\Cref{tab:corralign} measures how well students preserve the relational structure of their teachers using Frobenius distance \citep{singhal2001modern}, Pearson correlation \citep{benesty2009pearson}, and the structural similarity index (SSIM) \citep{wang2004image}. RRD maintains strong structural alignment, and combined with KD \citep{hinton2015distilling} attains the best overall correspondence. KD alone slightly leads standalone RRD on some metrics due to its direct logit matching, but combining RRD's global relational cues with KD's soft-target supervision produces the most faithful alignment.

\subsubsection{Semantic structure preservation.}

\Cref{tab:semantic} assesses the semantic organization of learned features via intra-class compactness, inter-class separation, normalized mutual information (NMI) \citep{vinh2010information}, and retrieval precision (mAP@5) \citep{zhao2022dkd}. RRD improves semantic clustering over CRD \citep{tian2022crd}, confirming that the \textit{information bottleneck} induced by asymmetric temperatures filters noise while retaining essential relational structure, yielding more coherent embedding spaces where similar classes remain proximally organized.

\subsection{Ablation Study}
\label{sec:ablations}

We ablate RRD on CIFAR-100 using a WRN-40-2 teacher and a WRN-16-2 student, repeating each experiment three times for consistency.

\subsubsection{Temperature Parameters}

\Cref{fig:ablation_temp} shows the best performance at $\tau_s = 0.1$ and $\tau_t = 0.02$. Accuracy degrades as $\tau_t$ grows and drops sharply once $\tau_t > \tau_s$, showing that an overly soft teacher harms distillation, consistent with our information-bottleneck interpretation. Excessively soft student distributions ($\tau_s = 1.0$) consistently underperform. The limit $\tau_t \to 0$ recovers an $\textit{argmax}$ that produces a one-hot target, whereas larger $\tau_t$ yields softer distributions that weaken teacher-student alignment.

\subsubsection{Memory Size}

\Cref{fig:ablation_memory} shows that accuracy improves with memory size for both RRD and CRD (\textit{repr.}), plateauing around $K = 16384$ with minimal gains beyond it. Separate memory banks for teacher and student would fail, since the KL divergence would then align distributions over different feature sets without direct feature matching. Storing student features in the bank would also create unstable targets, as student representations evolve continuously during training and undermine consistent supervision.

\subsubsection{Loss Weighting}

\Cref{fig:ablation_beta} shows that coefficients $\beta$ between $0.5$ and $1.5$ work best, matching the findings of \citet{tian2022crd}. Although we typically fix $\lambda = 1$, \Cref{fig:ablation_lambda} illustrates the effect of varying it.

\section{Conclusion}
\label{sec:conclusion}

We introduced a distillation framework that transfers knowledge by aligning relational similarity structures between teacher and student through a controlled \textit{information bottleneck}. Experiments across image classification, transfer learning, and object detection demonstrate that RRD consistently outperforms conventional and contrastive distillation approaches across diverse datasets and architectures. By emphasizing relative relationships rather than absolute feature matching, RRD preserves the structural integrity of learned representations, yielding robust, generalizable models that capture the teacher's relational geometry while delivering superior downstream performance and representational consistency across tasks and scales.

\subsection*{Scope and Future Work}

The primary goal of this work is to demonstrate that relaxing strict contrastive objectives into soft relational alignment yields a more effective distillation signal. While we focus on convolutional architectures to ensure fair comparison with prior contrastive distillation methods, extending RRD to transformer-based architectures represents a promising direction for future work.

\section*{Acknowledgements}
We acknowledge the computational resources and support provided by the Imperial College Research Computing Service (\url{http://doi.org/10.14469/hpc/2232}), which enabled our experiments.

{
    \small
    \bibliographystyle{ieeenat_fullname}
    \bibliography{main}
}

\end{document}